\documentclass[sigconf]{acmart}

\AtBeginDocument{%
  }
\usepackage{amsmath}
\usepackage{hyperref}
\usepackage{subcaption}
\usepackage{caption}
\usepackage{booktabs}
\usepackage{multirow}
\usepackage{geometry}
\setcopyright{acmlicensed}
\copyrightyear{2026}
\acmYear{2026}
\acmDOI{XXXXXXX.XXXXXXX}
\acmConference[BUILDSYS'26]{Make sure to enter the correct
  conference title from your rights confirmation email}{June 22--25,
  2026}{Banff, Alberta, Canada}
\acmISBN{978-1-4503-XXXX-X/2018/06}

\begin{document}

\title{Toward a foundational thermal model for residential buildings}


\author{Ting-Yu Dai, Kingsley Nweye}
\email{{funnyengineer,nweye}@utexas.edu}
\affiliation{%
  \institution{The University of Texas at Austin}
  \city{Austin}
  \state{Texas}
  \country{USA}
}

\author{Dev Niyogi}
\email{dev.niyogi@jsg.utexas.edu}
\affiliation{%
  \institution{The University of Texas at Austin}
  \city{Austin}
  \state{Texas}
  \country{USA}
}

\author{Zoltan Nagy}
\email{z.nagy@tue.nl}
\affiliation{%
  \institution{Eindhoven University of Technology}
  \state{Eindhoven}
  \country{The Netherlands}
}

%% Abstract
\begin{abstract}
The building energy community lacks a foundational thermal model, i.e., a single pretrained model capable of generalizing across diverse buildings, climates, and control strategies without building-specific calibration. Achieving this vision requires architectural principles that capture universal thermal dynamics rather than memorizing building-specific patterns. We take a step toward this goal by presenting a physics-informed transformer architecture that embeds domain knowledge, e.g., derivative enrichment and Euler-based numerical integration, into a decoder-only framework. We incorporate static building features extracted from simulation models and employ Rotary Position Embedding attention to capture temporal dependencies. Evaluated on the CityLearn dataset spanning 247 residential buildings across three climate zones, our model achieves one-step prediction accuracy (RMSE of 0.30°C in Texas, 0.29°C in Vermont) while outperforming both traditional baselines and fine-tuned Time-Series Foundation Models. We also demonstrate zero-shot transferability: models trained on as few as two buildings generalize to unseen buildings and climate zones without fine-tuning. Despite the limitation of simulated residential buildings, our results establish physics-informed architectural principles as a promising foundation for universal building thermal models.
\end{abstract}


%% Keywords
\maketitle



\section{Introduction}
% Illstrate the challenges for building energy consumption, focus on big picture. climate change global warming, final conclude to the renewable energy source
\begin{figure*}[tbh]
    \centering
    \includegraphics[width=\textwidth]{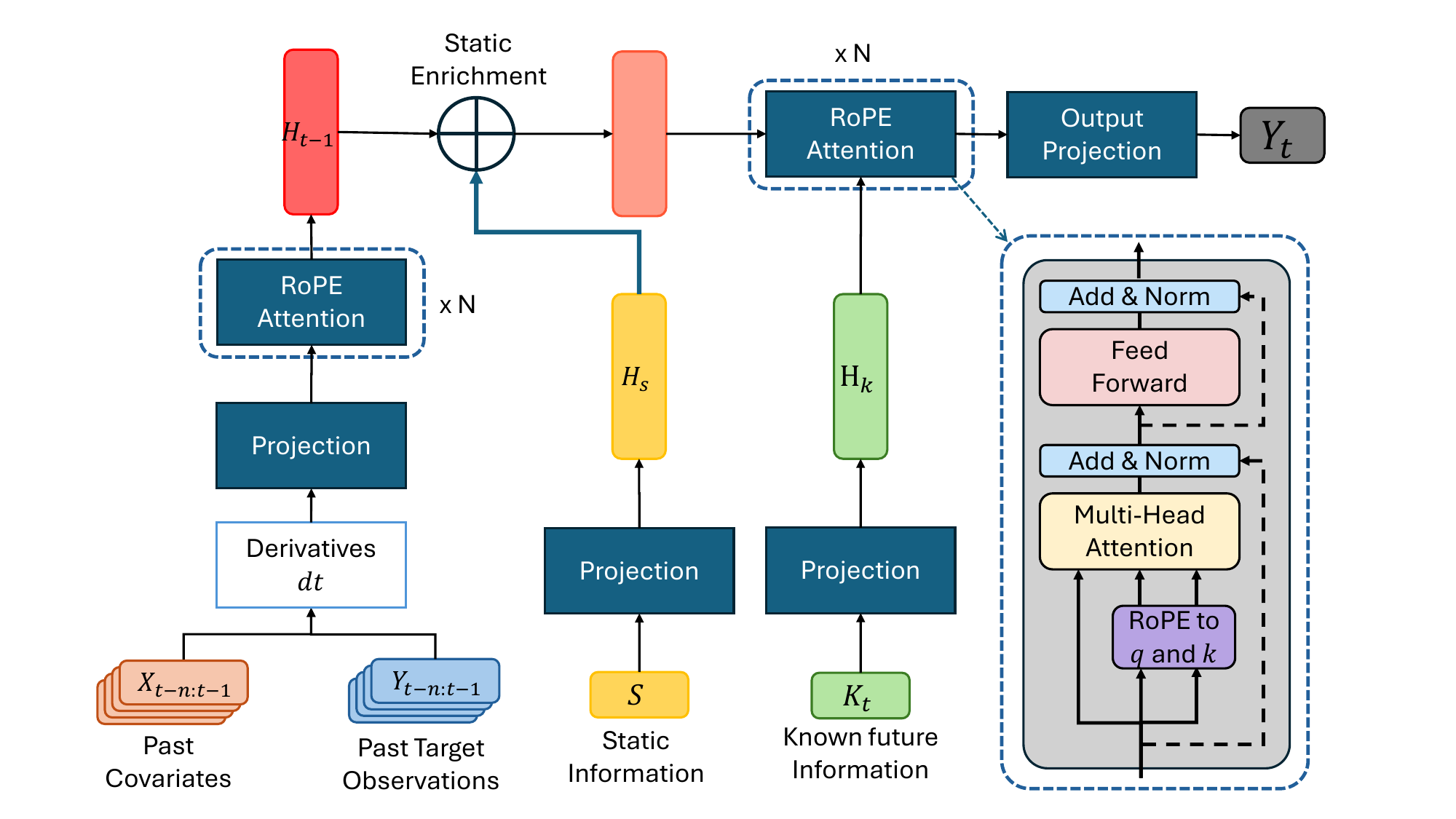}
    \caption{Physics Transformer architecture. Past observations and their derivatives are encoded by RoPE attention layers, then enriched with static building features ($H_S$). A second attention stack integrates future-known covariates ($H_k$) before predicting the temperature change $\Delta T_t$ The inset (right) shows the transformer block structure with RoPE applied to queries and keys.}
    \label{fig:method}
\end{figure*}

Buildings account for 40\% of global energy consumption, making them central to decarbonization efforts~\cite{un2025report}. Their inherent thermal storage and controllable loads enable demand flexibility, but realizing this potential requires accurate thermal models for advanced control strategies such as Model Predictive Control (MPC) and Reinforcement Learning (RL).

The successful realization of this potential, however, is essentially predicated on the deployment of robust and adaptive control strategies, Model Predictive Control (MPC) and Reinforcement Learning (RL). The efficacy of these advanced controllers is highly correlated to the availability of accurate, reliable, and computationally tractable building energy models (BEMs) that capture the thermal dynamics of buildings. Without a precise understanding and prediction of a buildings' thermal behavior, optimizing  energy operations becomes an intractable challenge, enabling faster ways to simulate thermal dynamics, potentially through data-driven approaches.

Despite significant research, developing control-oriented thermal models encounters persistent hurdles. Physics-based models require building-specific calibration; data-driven models need large datasets and generalize poorly; grey-box RC models struggle with nonlinearities. Recent Time-Series Foundation Models (TSFMs) exhibit limited generalizability for building energy tasks~\cite{mulayim2024time,park2025probabilistic}, and their scale is mismatched with control latency requirements. Prior work on LSTM-based models~\cite{jeon2021lstm,jang2022prediction} and few-shot transfer~\cite{tang2023privacy,lu2023multi} demonstrates potential but lacks unified standards for cross-building generalization. Thus, our driving research question is: \emph{how can we build a transferable thermal model capable of zero-shot generalization to unseen buildings?}

The success of foundation models in natural language processing and computer vision has sparked interest in developing analogous models for engineering domains~\cite{das2024decoder,wiesner2025towards}. A foundational thermal model for buildings would ideally exhibit three properties: (1) zero-shot generalization to unseen buildings without fine-tuning, (2) cross-domain transfer across building types, climates, and operational contexts, and (3) sample efficiency requiring minimal data from new buildings for adaptation. Achieving these properties for buildings presents unique challenges absent in language or vision domains. Buildings are governed by well-understood physics (heat transfer, thermodynamics), but exhibit  heterogeneity in geometry, materials, systems, and occupancy. Unlike images or text, building time series are inherently multi-modal, coupling weather, occupant behavior, and HVAC operation. Also, obtaining labeled training data requires either expensive real-world instrumentation or computationally intensive physics-based simulation.

We hypothesize that foundational thermal models require physics-informed architectural priors, not only large amounts of data. While general Time-Series Foundation Models (TSFMs) achieve broad coverage through massive pre-training, they lack inductive biases aligned with physics. We test this hypothesis by integrating Euler-based residual prediction (learning temperature changes rather than absolute values) and derivative feature augmentation (explicitly encoding rate-of-change dynamics) into a decoder-only transformer with Rotary Position Embedding (RoPE) attention. We do not claim to have achieved a complete foundational model; rather, we establish architectural principles and provide empirical evidence for zero-shot transfer that we believe are necessary steps toward this goal. Our specific contributions are:
\begin{itemize}
\item Physics-Informed Architectural Principles: We identify and validate two inductive biases, i.e., derivative enrichment and Euler-based residual prediction, that embed heat transfer physics into neural architectures. We demonstrate that these principles enable data-efficient transfer learning. %Our model achieves competitive accuracy with ~100K parameters, compared to 200M parameters in general-purpose TSFMs.
\item Empirical Evidence for Zero-Shot Transfer: We provide the first systematic evaluation of thermal model transferability across both buildings (within-climate) and climate zones (cross-climate). %Models trained on as few as two buildings achieve strong generalization to 100 unseen buildings; models trained on 32 buildings across three climates transfer to all 247 buildings without fine-tuning (RMSE < 0.50°C). This demonstrates that the zero-shot generalization property of foundational models is achievable for building thermal dynamics.
\item Benchmarking the Foundation Model Paradigm: We compare physics-informed design against the prevailing foundation model approach (scale + generic architecture). %Our results show that domain-specific architectural choices outperform fine-tuned TimesFM, suggesting that foundational models for scientific domains may require different strategies than those successful in NLP and vision.
\end{itemize}

%Unlike generic TSFMs that rely on their massive parameters, our architecture utilizes physics-informed design, such as derivative enrichment and Euler integration, to achieve generalization with a fraction of the parameters. We hope this work bridges the gap between large TSFMs and building energy modeling, surrogate models, and provides a more scalable and simpler tool for building energy researchers and practitioners.

The next Section states the problem and the introduces our proposed Physics Transformer architecture. Then, Section~\ref{s:experiment} details the experimental setup. Section~\ref{s:results} presents the results and discusses their implications, while Section~\ref{s:conclusion} concludes the paper.

\section{Methodology}
\subsection{Problem Statement}
Our goal is to build a building thermal model that forecasts the indoor temperature $T$ at target timestamp $t$ of building $j$:
\begin{equation}
    T_t = f\left(T_{t-n:t-1}, W_{t-n:t-1}, S_j, K_t\right)
    \label{equ:prob}
\end{equation}
The model receives the past $n$ time points as a time series, which contains past target observation $T_{t-n:t-1}$ and other past-observed covariates $W_{t-n:t-1}$, static covariates $S$ from building $j$ , and future-known covariates $K_t$. The key challenge is learning thermal dynamics that generalize to unseen buildings and climates.

\subsection{Proposed Physics Transformer Architecture}
Figure~\ref{fig:method} illustrates the proposed architecture, which combines a decoder-only transformer backbone~\cite{das2024decoder} with static feature enrichment~\cite{lim2021temporal}. Past observations and their temporal derivatives are embedded into a 64-dimensional hidden space, then encoded by $N$ RoPE attention layers to produce a temporal representation ($H_{t-1}$). Static building features ($H_S$) enrich this representation before a second attention stack integrates future-known covariates ($H_k$). The output layer predicts the temperature change $\Delta T_t$ rather than the absolute temperature. We now describe each component:

%We are inspired by TimesFM~\cite{das2024decoder} to create a decoder-only based model with static enrichment from Temporal Fusion Transformer \cite{lim2021temporal}. Our model takes as inputs past observed target and covariates, static building features, and future known covariates. The past observed target and covariates have been processed to compute the temporal derivatives, and then all past observations, static information, and future-known variables are embedded into hidden space though a linear transformation where hidden dimension is set to 64. The embedded historical features are being encoded by $N$ layers of Rotary Position Embedding (RoPE) Attention to produce a latent historical representation ($H_{t-1}$). This representation is then enriched with static building features ($H_s$). The resulting enriched context is fed into the second attention stack, which integrates projected known future information ($H_k$) to condition the forecast. Finally, a small feedforward network is used to predict the change in indoor temperature ($\Delta T_t$) for the next time step. Specifically:

\textbf{Single Step Prediction}
Our model predicts only the next timestep temperature, following the autoregressive paradigm where each prediction conditions on ground-truth history. While multi-horizon output heads can improve forecasting efficiency~\cite{das2024decoder}, they require assuming known future covariates (e.g., HVAC loads), which is typically unavailable in realistic control scenarios where the controller determines future actions. 

%[REWORK]We follow the state machine fashion to only predict the target indoor temperature of next timestep. Although \cite{das2024decoder} mentioned that produce multiple future timesteps at once yields better result, other input variables might provide different while performing thermal dynamics during later training policies. Therefore, predicting future variates while assuming known further dynamics, like building loads is inapplicable in our study. 

\textbf{\emph{Decoder-only} Architecture}
We adopt a decoder-only transformer architecture, which is inherently autoregressive: each prediction depends on the preceding sequence. This aligns naturally with thermal forecasting, where the next temperature depends on historical states. The architecture has proven effective for time series foundation models~\cite{das2024decoder}.

%The decoder-only model has been widely adopted for current large language model architectures. A decoder-only model is a variant of the transformer architecture that consists solely of a stacked decoder layer. This design is essentially autoregressive, meaning the model focuses on generating next token at a time, where each new token is predicted based on previously generated sequences. This is aligned with our task to predict the indoor temperature at the next timestep. Such models are highly effective for generative tasks. Pre-trained on a simple, self-supervised objective of next target prediction, decoder-only models, such as the recent time series foundation model~\cite{das2024decoder}, have demonstrated powerful capabilities in time series forecasting.

\textbf{Static Enrichment}
The difference in thermal mass, insulation, and geometry of buildings fundamentally shapes their thermal response. We incorporate these time-invariant characteristics through static enrichment, inspired by the Temporal Fusion Transformer~\cite{lim2021temporal}. Unlike TFT's repeated static injection for interpretability, we apply a single additive fusion post-temporal encoding for computational efficiency. Specifically, static building features, $H_s$, (see Section~\ref{ss:staticParams}) are projected to the hidden dimension via a linear layer, then added to the encoded temporal representation ($H_{T-1}$). This allows the model to modulate its predictions based on building-specific physics, e.g., predicting slower temperature changes for well-insulated buildings—without requiring these relationships to be learned purely from temporal patterns.

\textbf{Derivative Enrichment}
Inspired by neural operators~\cite{wiesner2025towards,kovachki2023neural}, we introduce an inductive bias centered on differential operators. To make model explicitly capture the rate-of-change dynamics in heat transfer process, we augment the input space with the first-order temporal derivative $\frac{dX}{dt}$. The approximated derivative is computed by second-order centered difference schemes:
 \begin{equation}
    f'(t) = \left[f(t+ \Delta t) - f(t-\Delta t)\right] / (2\Delta t)
    \label{equ:deriv}
 \end{equation}

\textbf{Numerical Integration}
Building thermal dynamics are governed by heat balance equations: the rate of change of the indoor air temperature depends on heat gains (solar, internal, HVAC power) and losses (conduction, ventilation, infiltration). Inspired by residual learning in weather forecasting task \cite{price2025probabilistic, mardani2025residual} and neural operators \cite{wiesner2025towards}, we embed physical structure by training the model to predict temperature changes rather than absolute values:
\begin{equation}
    T_t = T_{t-1} + f(T_{t-n:t-1}, W_{t-n:t-1}, S_j, K_t)
    \label{equ:num_int}
\end{equation}
This formulation is equivalent to explicit Euler integration of an ordinary differential equation (ODE). The key benefit for generalization is that the model learns thermal response patterns (e.g., "this heat input causes 0.5°C/hour rise") rather than absolute temperature ranges (e.g., "this building operates at 21-23°C"). Since thermal response patterns are more consistent across buildings than absolute setpoints, this inductive bias directly supports zero-shot transfer. Additionally, constraining outputs to temperature changes (typically $<1^\circ\text{C/hour}$) provides implicit regularization compared to predicting absolute temperatures spanning $15-30^\circ$C.

% %Inspired by residual learning in weather forecasting task \cite{price2025probabilistic, mardani2025residual} and neural operators \cite{wiesner2025towards}, we formulate the prediction task as learning the differential dynamics of the system. Rather than predicting the absolute indoor temperature $T_t$ directly, the model learns the residual dynamics $\Delta T_t$. The final target is reconstructed via:

% \begin{equation}
%     T_t = T_{t-1} + f(T_{t-n:t-1}, W_{t-n:t-1}, S_j, K_t)
%     \label{equ:num_int}
% \end{equation}

%  [REWORK] This formulation is mathematically equivalent to a one-step explicit Euler integration, widely used in numerical solvers for Ordinary Differential Equations (ODEs). As the network learns the gradient of the temperature, we align the learning objective with the underlying physics of heat transfer. This mechanism aligns with the \emph{Learning to Compare} paradigm in few-shot learning, where models achieve generalization by learning a relational metric between samples rather than absolute class identities. This allows the model to apply heat transfer knowledge to unseen buildings instead of remembering certain dynamics supplied in the training data.

\textbf{RoPE Attention} We use Rotary Position Embedding (RoPE)~\cite{su2024roformer} to encode temporal position. Unlike absolute positional encodings, RoPE applies rotation matrices to query and key vectors, encoding relative position through the angle between vectors. This enables the model to generalize to sequence positions not seen during training and captures the intuition that thermal dynamics depend on relative time gaps rather than absolute timestamps. Each attention block follows standard transformer structure: multi-head attention with RoPE-transformed queries and keys, followed by residual connections, layer normalization, and a feed-forward network (Fig.~\ref{fig:method}).

%[REWORK] We use a Rotary Position Embedding (RoPE) attention~\cite{su2024roformer}, which instead of adding static or learnable positional information, rotates the query and key vector before computing the scaled dot product attention. RoPE attention has been widely applied in recent large language models such as LLaMA and GPT-OSS. As detailed in Fig.~\ref{fig:method}, each RoPE attention block follows a standard Transformer layer structure—comprising multi-head attention, residual connections with layer normalization, and a feed-forward network, with the key distinction that RoPE is applied to the query ($q$) and key ($k$) vectors to inject relative positional information.

\textbf{Training Details}
The model is trained with AdamW \cite{kingma2014adam} (learning rate 0.003125, weight decay 0.01) and reduce the learning rate by $0.25\times$ after 3 epochs without validation improvement.  The batch size is 4096, and the max epoch is set to 400. Early stopping is employed based on the failure to improve validation loss in 10 consecutive epochs to prevent overfitting. Training uses a single NVIDIA A100 GPU (40GB) with PyTorch Lightning. The loss function is Mean Squared Error (MSE).
\section{Experimental Setup}
\label{s:experiment}

\subsection{Data Preparation}
\label{sec:data_acq}
\label{ss:staticParams}

We use the CityLearn dataset~\cite{nweye2025citylearn}, containing simulated thermal responses for 247 residential buildings: 100 in Travis County, TX (climate zone 2A), 100 in Alameda County, CA (zone 3C), and 47 in Chittenden County, VT (zone 6A). Buildings are modeled using EnergyPlus~\cite{crawley2001energyplus} with representative archetypes from RESSTOCK~\cite{present2024resstock}. Each building is simulated under five HVAC control strategies to generate diverse temperature profiles: (1) default mechanical HVAC as reference; (2) ideal loads meeting all heating/cooling demand; (3) generic equipment validation against ideal loads; (4) free-floating with no conditioning; and (5) stochastic load variation where the ideal load is scaled by factors 0.3–1.7 with 60\% probability of deviation~\cite{deltetto2020data}. This diversity ensures the model encounters varied thermal dynamics during training.

We extract static building parameters from EnergyPlus IDF files (using the \texttt{eppy} and \texttt{opyplus} packages) to condition the model on time-invariant physics. The features include: total floor area, aspect ratio (length/width), window-to-wall ratio (WWR), area-weighted average R-values for walls and roof, and internal gain density (W/m² from occupants, lighting, and equipment). These parameters capture the thermal mass, insulation quality, and heat gain characteristics that govern a building's thermal response independent of weather or HVAC operation.

% A key feature in our model is that we not only receive the input of past dynamics but also the static, physics-based information from the EnergyPlus IDF files. This step enriches the dataset for the proposed model with building geometry and system level characteristics. We apply the \texttt{eppy} and \texttt{opyplus} package to parse each IDF file to compute a set of static parameters. The key extracted features include:
% \begin{itemize}
%     \item Total Floor Area: Sum of floor areas of all thermal zones.
%     \item Aspect Ratio: The ratio of the building's longer dimension to its shorter dimension, based on the building's footprint.
%     \item Window-to-Wall Ratio (WWR): The ratio of the total window area to the total gross exterior wall area.
%     \item Area-Weighted Average R-values: The average thermal resistance for the building's exterior walls and roof, weighted by the area of each surface. The R-value for each construction assembly is calculated by summing the thermal resistances of its constituent material layers. The resistance of each layer is derived from its thickness and thermal conductivity, or taken directly if it is a NoMass or AirGap material.
%     \item Internal Gains: The sum of the design-level load densities (W/m$^2$) for occupants, lighting, and electrical equipment normalized by the total floor area.
% \end{itemize}

\subsection{Data Split}
\label{sec:data_split}
To evaluate generalization, we use the default mechanical HVAC strategy (mode 1) as a held-out test set, ensuring evaluation on unseen control dynamics. From the remaining data (modes 2–5), we randomly sample 10\% of each simulation's time series for validation; the rest is used for training. We denote results on the held-out HVAC strategy as \emph{Test} (unseen dynamics) and results on the 10\% sample as \emph{Validation} (seen dynamics, unseen timesteps). All 247 buildings appear in the test set; the validation set varies with the number of training buildings. We report Root Mean Squared Error (RMSE) and Mean Absolute Percentage Error (MAPE).

%Due to the aim of generalization ability in our proposed work, we evaluate our model in multiple situations. We first take the mechanical load e.g. the first mode that we generated in Sec. \ref{sec:data_acq} as the separate test set so that the model is evaluated on unseen building control strategies. Then, for the rest of the data, we randomly segment a certain portion of time series data from each simulation as the validation set (10\%) to establish a fair comparison across different dynamics we made in Sec. \ref{sec:data_acq}. The remaining data is used for training the model. We refer to the evaluation metric in the first setting as the \textbf{Test} since it's testing on unseen dynamics, and the second setting as \textbf{Validation} since it's validating on seen buildings and dynamics. We use Root Mean Squared Error (RMSE) and Mean Absolute Percentage Error (MAPE) as the main evaluation metrics. Note that all 247 buildings across three climate zones are included in test set, but the validation set only contains parts of the dataset so that fair evaluation could be made with unseen building models.

% To access the model's generalization to unseen buildings, further results are we also conduct a building-wise cross-validation. Specifically, we randomly select 20\% of buildings as the test set, and the remaining buildings are used for training and validation (with a 90-10 split). This approach ensures that the model is evaluated on entirely new building configurations that were not present during training.

\subsection{Baseline Models}
\label{sec:baseline_model}
To evaluate the performance of the proposed model, we compare it against four baselines: (1) \textbf{Individual LSTM}~\cite{nweye2025citylearn} per building, trained and tested on the same building with different HVAC strategies.
%as the baseline. The LSTM model takes in the same input features as our proposed model, including past observed indoor temperature, covariates, and future known covariates, but not incorporate static building features.%\footnote{The dynamic error is presented at \href{https://github.com/intelligent-environments-lab/CityLearn/tree/master/data/datasets}{https://github.com/intelligent-environments-lab/CityLearn/tree/master/data/datasets}} 
%They trained one LSTM model with one building at a time, and evaluated the model on unseen control strategies but for the same building. In contrast, our proposed model is trained on multiple buildings simultaneously to learn a universal thermal model that can generalize to unseen buildings and control strategies and evaluate across all buildings. 
%Additionally, we also implemented other baseline models that are trained on the same multiple buildings including LSTM, XGBoost~\cite{chen2016xgboost}, and TimesFM~\cite{das2024decoder}. For the LSTM model, We replace the RoPE attention blocks in our proposed model framework in Fig.~\ref{fig:method} with same number layers of stacked LSTM while keeping the rest of the architecture unchanged. %We use TimesFM for two comparison: 1.) This allows us to directly compare the impact of the attention mechanism versus recurrent layers within the same overall model structure.
(2) \textbf{Multi-building LSTM:} our architecture (Fig.~\ref{fig:method}) with RoPE attention replaced by stacked LSTM layers, trained on multiple buildings. (3) \textbf{XGBoost}~\cite{chen2016xgboost}: gradient-boosted trees trained on the same multi-building data. (4) \textbf{TimesFM}~\cite{das2024decoder}: a pretrained time-series foundation model, evaluated both zero-shot (200M parameters) and after fine-tuning on our dataset (109K params). All multi-building baselines use identical training data for fair comparison.

\begin{table*}[tbh]
\centering
\resizebox{\textwidth}{!}{%
\begin{tabular}{l cc c cc c cc}
\toprule
\multirow{2}{*}{\textbf{Model Type}} & \multicolumn{2}{c}{\textbf{Travis, TX}} & & \multicolumn{2}{c}{\textbf{Alameda, CA}} & & \multicolumn{2}{c}{\textbf{Chittenden, VT}} \\
\cmidrule{2-3} \cmidrule{5-6} \cmidrule{8-9}
 & \textbf{RMSE} ($^\circ$C) & \textbf{MAPE} (\%) & & \textbf{RMSE} ($^\circ$C) & \textbf{MAPE} (\%) & & \textbf{RMSE} ($^\circ$C) & \textbf{MAPE} (\%) \\
\midrule
Baseline (Individual LSTM) & 0.80 & 2.90 & & 0.52 & 2.00 & & 0.81 & 3.25 \\
LSTM                  & 0.59 & 1.7  & & 0.76 & 2.3  & & 0.51 & 1.75 \\
XGBoost               & 0.58 & 1.67 & & 0.56 & \textbf{1.97} & & 0.43 & 1.62 \\
TimesFM-200M w/o finetune              & 2.03 & 10.56 & & 2.97 & 12.21 & & 1.99 & 10.24 \\
TimesFM-109K retrained & 0.37 & 1.64 & & 0.50 & 2.50 & & 0.35 & 1.74 \\
\textbf{PhysicsTransformer} & \textbf{0.30} & \textbf{1.46} & & \textbf{0.48} & 2.37 & & \textbf{0.29} & \textbf{1.42} \\
\bottomrule
\end{tabular}%
}
\caption{Cross-regional performance on \textbf{Test Set} comparison of various models trained on 32 buildings from all three climates. All models except TSFM-200M are trained on equivalent datasets.}
\label{tab:model_comparison_all}
\end{table*}

\subsection{Evaluation Protocol}
We evaluate generalization in two settings. \emph{Single-climate experiments} train on 1--32 buildings from one region and test on all buildings in that region. Building selection is nested: the $n$-building training set is always a subset of the ($n+1$)-building set, ensuring performance changes reflect data diversity rather than building-specific effects. We distinguish \emph{target buildings} (seen during training) from \emph{non-target buildings} (unseen) to quantify generalization.
\emph{Cross-climate experiments} train on buildings drawn equally from all three regions (minimum 4 total, at least 1 per region) and evaluate on all 247 buildings. We also apply single-region models to other regions without fine-tuning to assess zero-shot climate transfer.

\section{Result}
\label{s:results}
\subsection{Overall Performance}
Table \ref{tab:model_comparison_all} compares the performance on the test set. We evaluate our proposed Physics Transformer (Cross-Region model) against the baselines described above. The Cross-Region model was trained on a diverse dataset of 32 buildings spanning all three climate zones, and the baselines (XGBoost, LSTM, TimesFM) were trained under identical multi-building conditions to ensure a fair evaluation.

The zero-shot TimesFM-200M, despite being pre-trained on massive and diverse time series data, fails catastrophically on building thermal prediction ($\text{RMSE} > 2^\circ\text{C}$). This demonstrates that general-purpose foundation models do not automatically transfer to building energy, likely because the domain lies outside their effective pre-training distribution. This is consistent with findings in~\cite{mulayim2024time} that observed that TSFMs often perform only marginally better than statistical baselines on unseen building datasets. Fine-tuning (TimesFM-109K retrained) substantially improves performance, suggesting that the transformer architecture itself is capable but requires domain-specific adaptation.

%Our method achieves the lowest RMSE in the challenging Travis, TX and Chittenden, VT regions ($0.30^\circ C$ and $0.29^\circ C$, respectively), indicating that our physics-informed architecture captures universal thermal dynamics more effectively than generic architectures. Notably, the comparison highlights the limitations of general-purpose Time-Series Foundation Models (TSFMs) in this domain. The zero-shot TimesFM-200M exhibits significant error ($RMSE > 2^\circ C$), confirming that generic foundation models struggle with the heterogeneous dynamics of building physics. This is consistent with findings by \citet{mulayim2024time}, who observed that TSFMs often perform only marginally better than statistical baselines on unseen building datasets. Consequently, while fine-tuning (TimesFM-109K retrained) drastically improves performance, it still fails to match the accuracy of our specialized, physics-informed approach.

\subsection{Single Climate Zone Generalization}
\label{sec:single_region}
\begin{figure*}[htb]
    \centering
    \begin{subfigure}{0.49\textwidth}
        \centering
        \includegraphics[width=\textwidth]{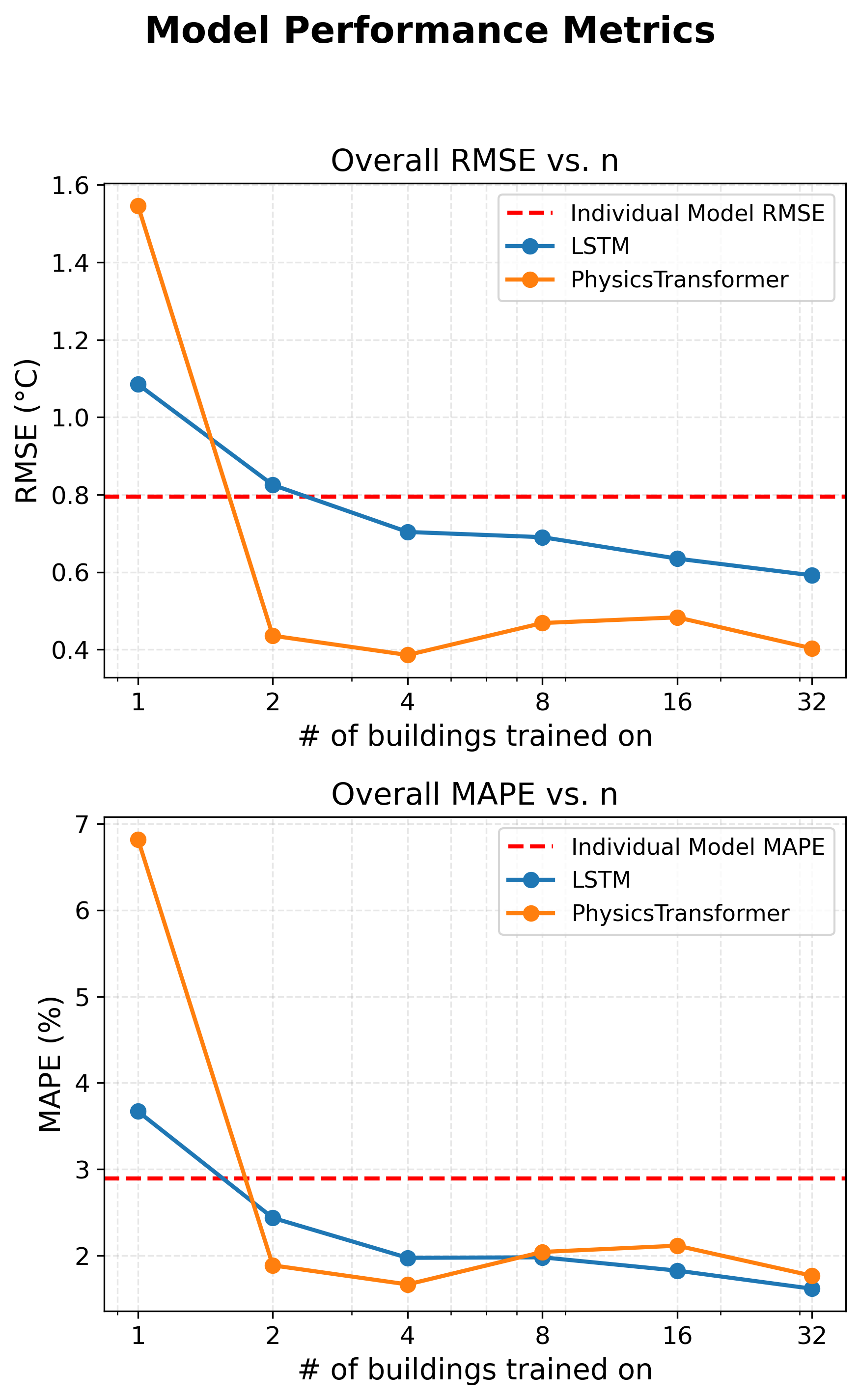}
        \caption{Overall model performance trained in different number of buildings in Travis County, Texas.}
        \label{fig:tx_single_overall}
    \end{subfigure}
    \hfill
    \begin{subfigure}{0.49\textwidth}
        \centering
        \includegraphics[width=\textwidth]{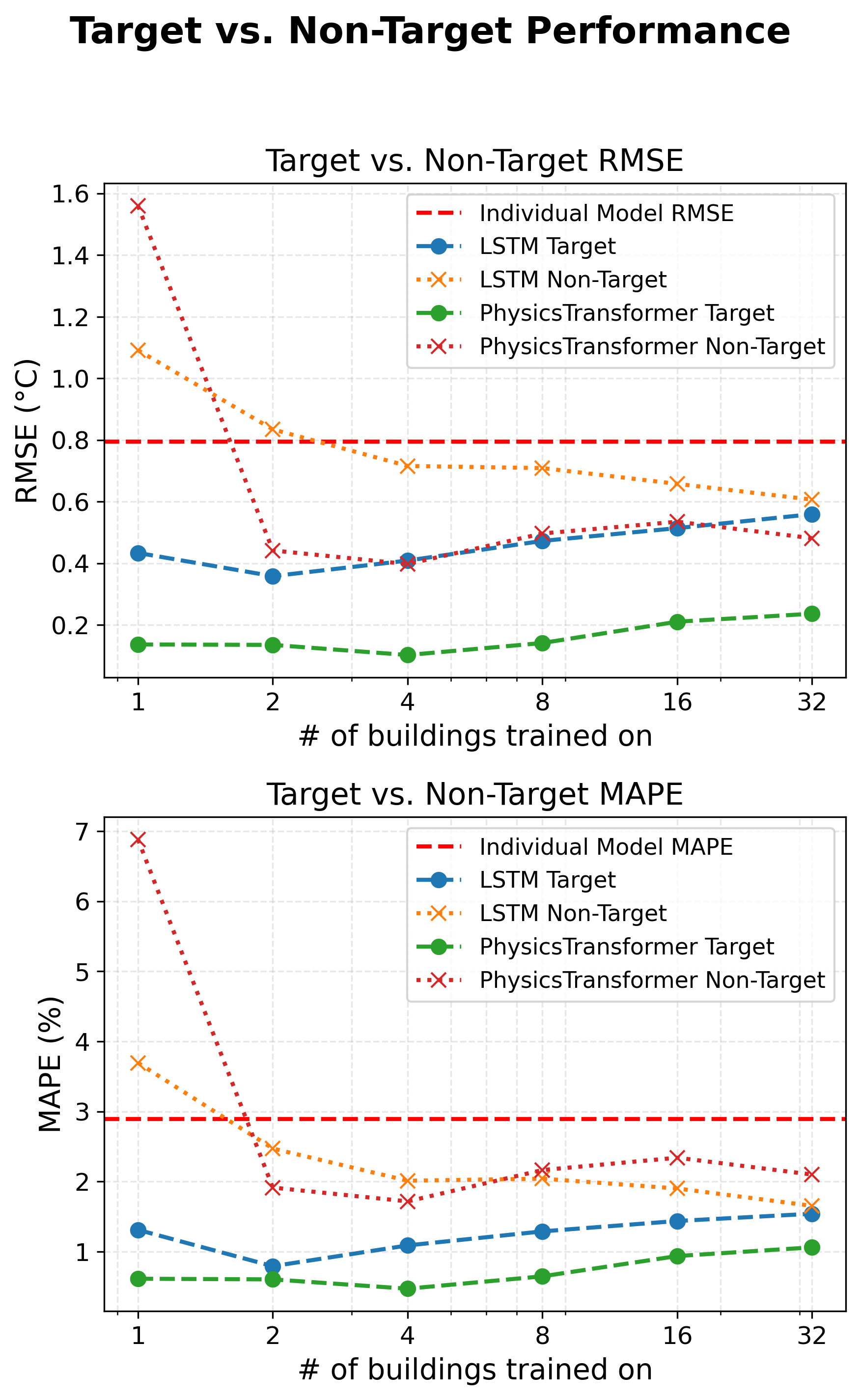}
        \caption{The comparison between training target buildings and non-target model performance in different number of buildings in Travis County, Texas.}
        \label{fig:tx_target}
    \end{subfigure}
    \caption{Texas region model performance comparison.}
    \label{fig:tx_combined}
\end{figure*}

Figure~\ref{fig:tx_combined} shows Texas results; California (Fig.~\ref{fig:ca_combined}) and Vermont (not shown) exhibit similar patterns. The Physics Transformer demonstrates strong data efficiency training on 2 buildings achieves performance comparable to LSTM trained on 32 buildings (Fig.~\ref{fig:tx_single_overall}), representing a 16 fold reduction in data requirements.

As training diversity increases, the gap between target and non-target buildings narrows (Fig.~\ref{fig:tx_target}), demonstrating generalization rather than memorization. The Physics Transformer shows greater improvement on non-target buildings than LSTM, indicating that physics-informed design better captures transferable dynamics.

\begin{figure*}[!ht]
    \centering
    \includegraphics[width=\textwidth]{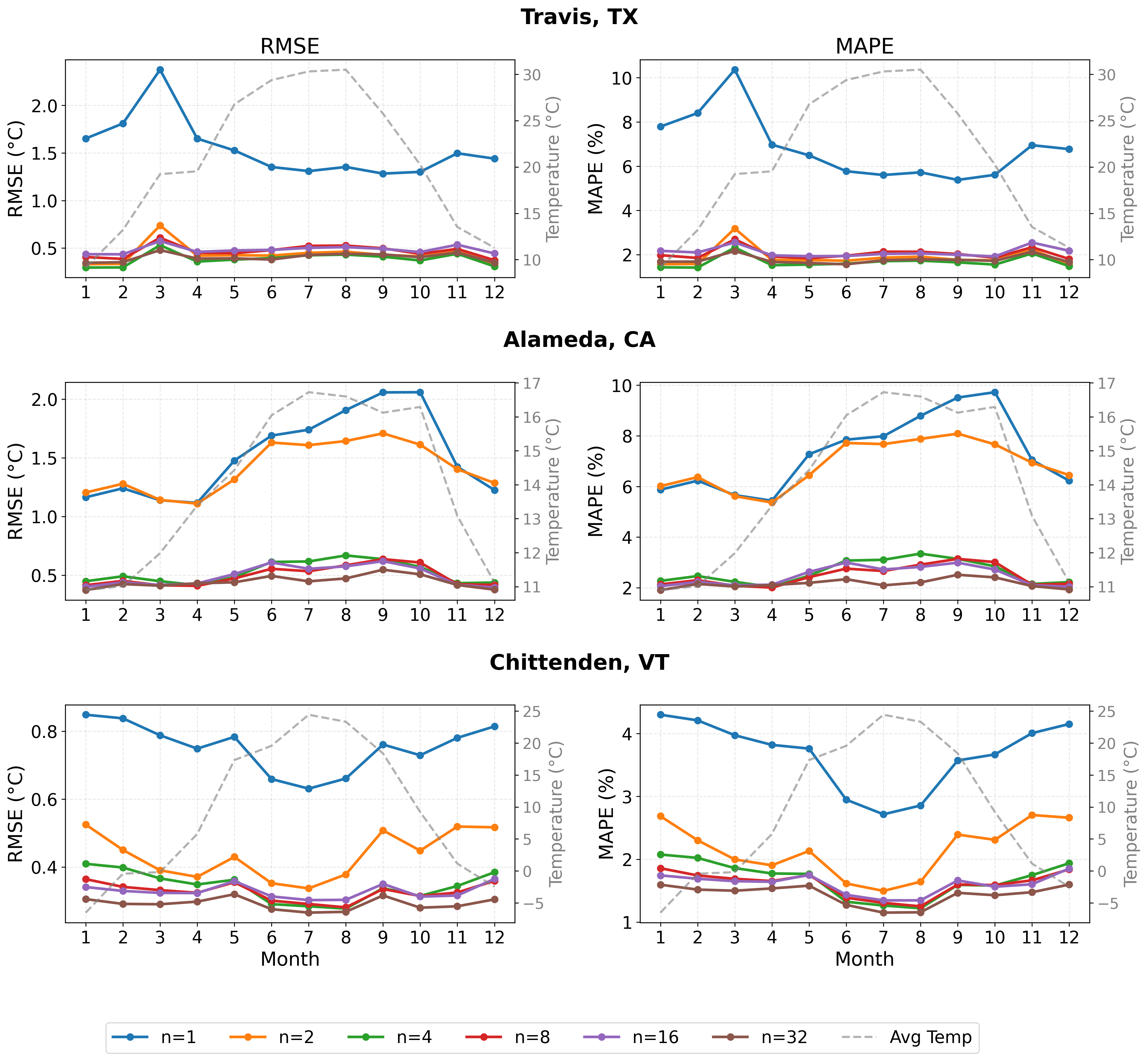}
    \caption{Monthly comparison between different regions and different number of buildings. The second axis shows the monthlyaverage outdoor temperature in each region.}
    \label{fig:monthly_single}
\end{figure*}

\begin{figure*}[htb]
    \centering
    \begin{subfigure}{0.49\textwidth}
        \centering
        \includegraphics[width=\textwidth]{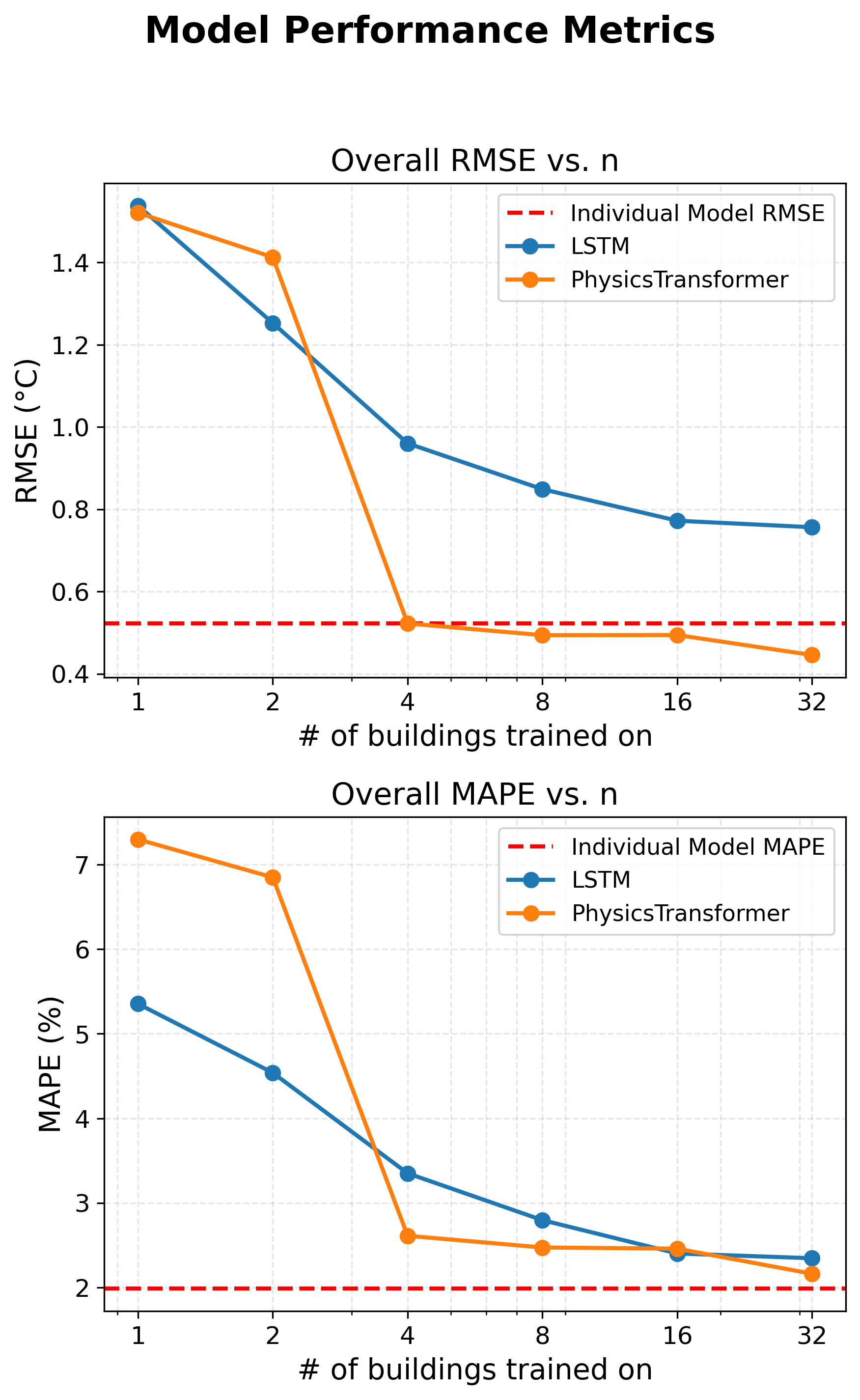}
        \caption{Overall model performance trained in different number of buildings in Alameda County, California.}
        \label{fig:ca_overall}
    \end{subfigure}
    \hfill
    \begin{subfigure}{0.49\textwidth}
        \centering
        \includegraphics[width=\textwidth]{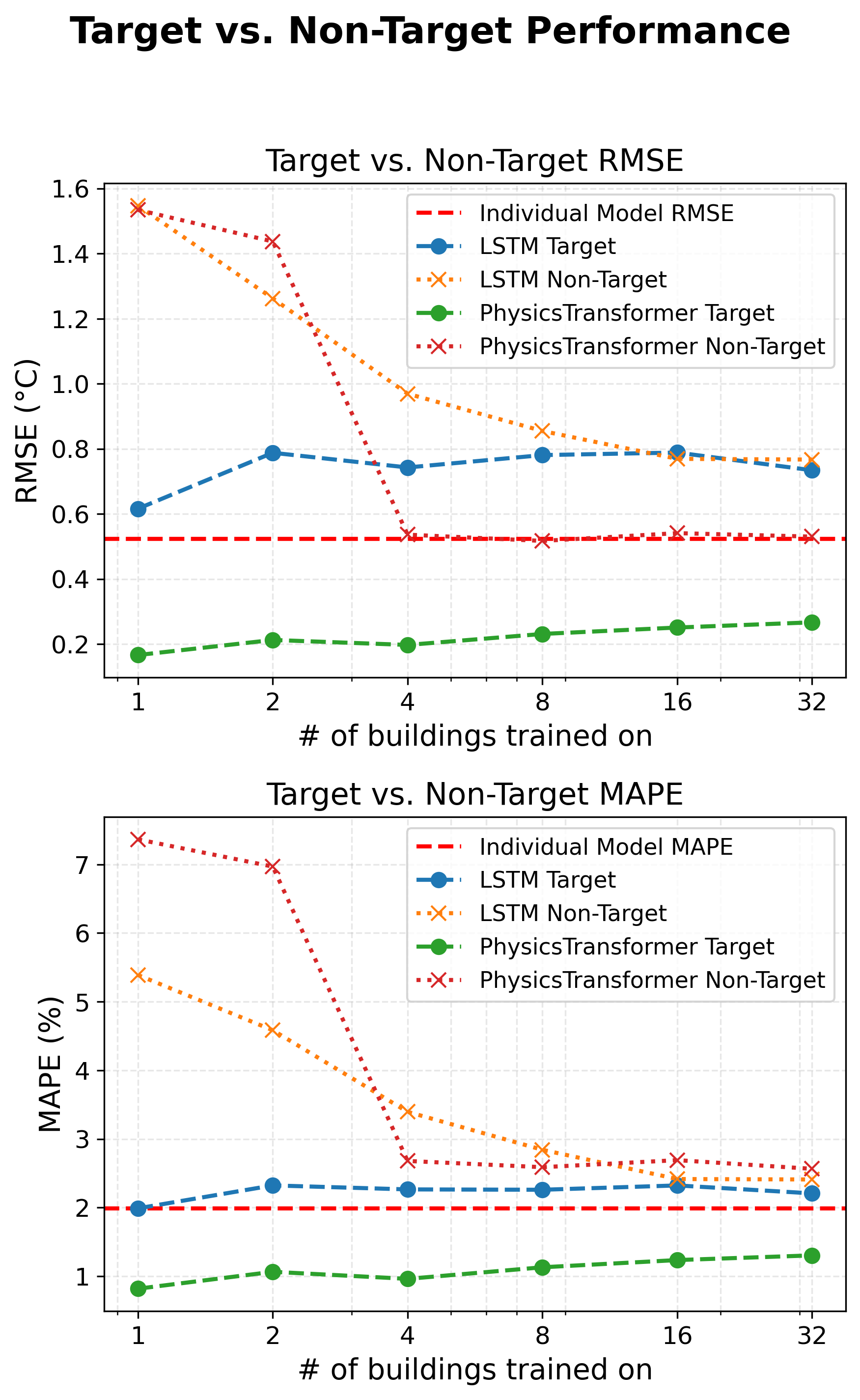}
        \caption{The comparison between training target buildings and non-target model performance in different number of buildings in Alameda County, California.}
        \label{fig:ca_target}
    \end{subfigure}
    \caption{California region model performance comparison.}
    \label{fig:ca_combined}
\end{figure*}

Figure~\ref{fig:monthly_single} shows the seasonal variation in performance. Errors increase during transition months (March in TX, May/September in VT) when weather patterns shift rapidly. Table~\ref{tab:single_region} reports validation set RMSE, which increases slightly with more training buildings as the validation set becomes more representative of the building population.

\begin{table}
    \centering
    \begin{tabular}{lccc}
        \toprule
        RMSE (°C) & Travis, TX & Alemeda, CA & Chittenden, VT  \\
        \midrule
        1 Building & 0.079 & 0.104 &  0.115 \\
        2 Buildings & 0.114 & 0.177 & 0.158  \\
        4 Buildings & 0.095 & 0.152 & 0.166 \\
        8 Buildings & 0.114 & 0.207 & 0.184 \\
        16 Buildings & 0.223 & 0.276 & 0.227 \\
        32 Buildings & 0.275 & 0.348 & 0.248 \\
        \bottomrule
    \end{tabular}
    \caption{Performance comparison on seen HVAC control strategies (Validation Set) in single region training.}
    \label{tab:single_region}
\end{table}

% CA IMAGE

\subsection{Cross-Climate Zone Generalization}
\begin{figure*}[htb]
    \centering
    \includegraphics[width=\textwidth]{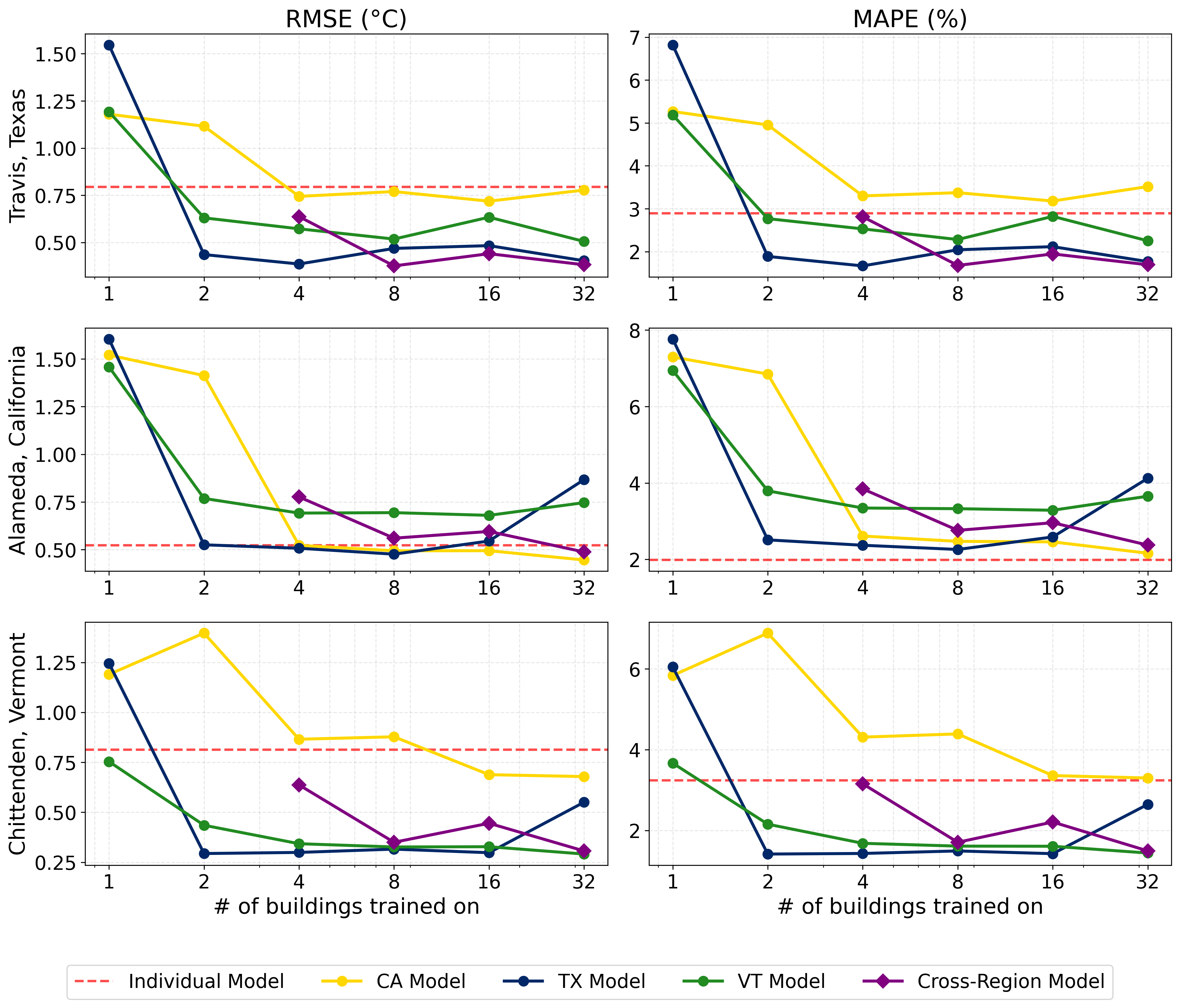}
    \caption{Cross region comparison between the multi-region trained model and single region model.}
    \label{fig:cross}
\end{figure*}

Figure~\ref{fig:cross} compares multi-region models against single-region models applied across climates. Multi-region models achieve performance comparable to, and sometimes exceeding, single-region models in their home regions.

Single-region models demonstrate zero-shot transfer: Texas-trained models achieve 0.48–0.55°C RMSE on California and Vermont without fine-tuning. Transfer quality varies by source region: Texas and Vermont models transfer effectively to other climates, while California models show degraded cross-climate performance. The 32-building Texas model also shows reduced transfer, suggesting potential overfitting at larger training scales.

%We now evaluate the model's performance when trained across multiple climate zones. We pick buildings from each region to have the same amount of training data for a fair comparison. Since at least one building should be extracted from each climate zone, the minimum number of multi-region model is 4 (see Fig.~\ref{fig:cross}). The test set remains the same as in the previous section to evaluate the model's generalization ability on unseen buildings. We also evaluate the trained single region model to the other regions to examine how it generalizes across different climate zones without any fine-tuning. Fig.~\ref{fig:cross} shows the comparison between the multi-region trained model and the single region model applied to other regions with the target model on target region as well. 

\subsection{Qualitative Analysis}
\begin{figure*}[tbh]
    \centering    \includegraphics[width=.8\textwidth]{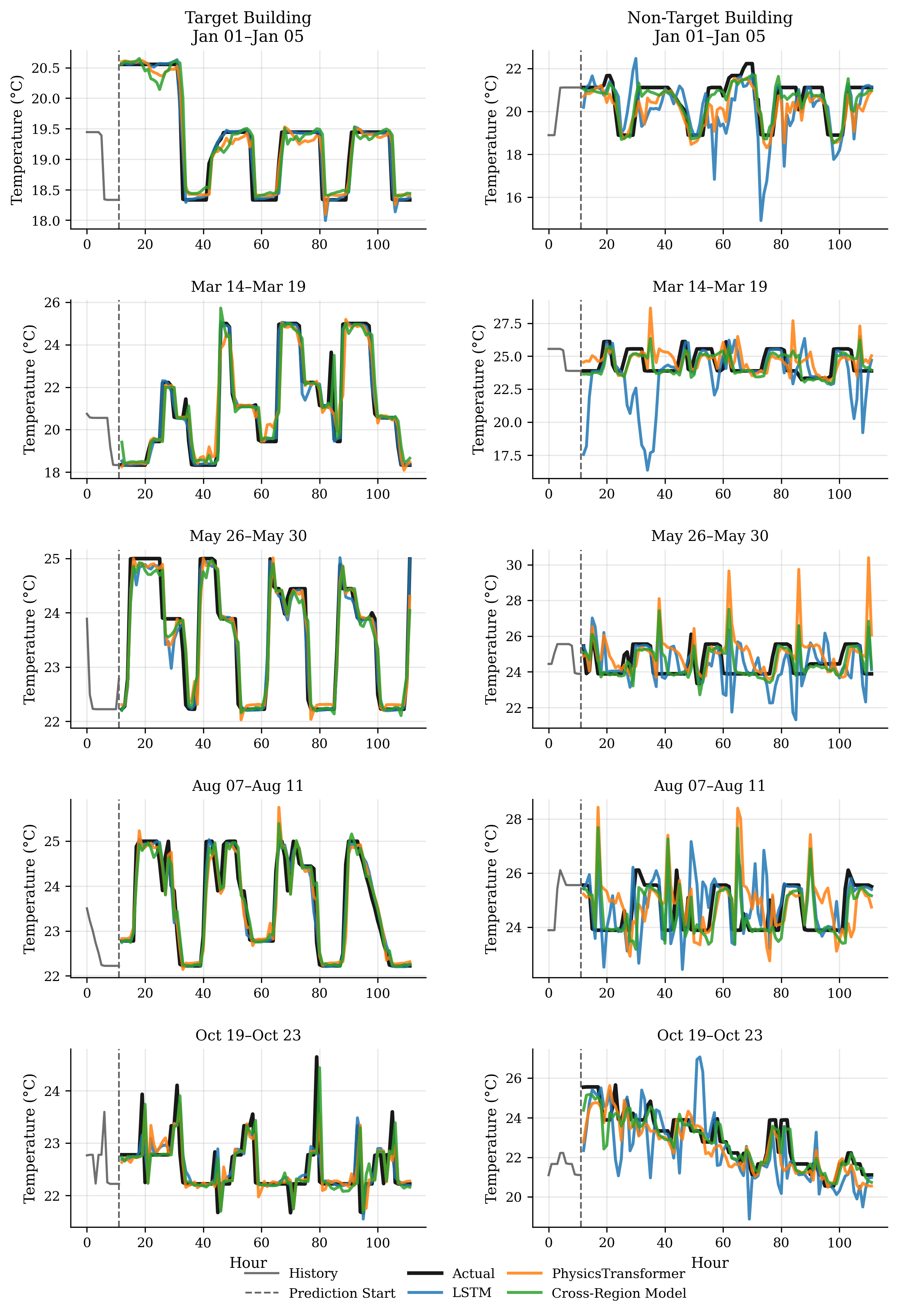}
    \caption{Comparisons of one-step prediction in January, March, May, August, and October in Travis County, Texas.}
    \label{fig:actual_tx}
\end{figure*}

Figure~\ref{fig:actual_tx} presents representative one-step predictions for Texas, for both a target building, and a non-target building. Each row shows a different season to illustrate performance across varying thermal conditions. On the target building, all models track the ground truth closely with minimal visible error, confirming the low RMSE values reported in Table~\ref{tab:single_region}.

On the non-target building, differences between models become apparent. The Physics Transformer produces smoother predictions that follow the overall thermal trajectory, while LSTM shows higher variance with more erratic fluctuations. Both models capture the diurnal cycling pattern but occasionally miss rapid temperature changes, particularly during transition months (March, October) when weather variability is highest. This demonstrates qualitatively that the Physics Transformer generalizes more effectively to unseen buildings.

\section{Discussion}
\subsection{Assessment Against Foundational Model Criteria}
We evaluate our model against the three foundational model properties defined in in the Introduction.
\begin{enumerate}
    \item\emph{Zero-shot generalization:} Our model demonstrates strong zero-shot transfer within the evaluated domain. Models trained on Texas buildings achieve RMSE of 0.48-0.55°C on California and Vermont buildings without any fine-tuning (Figure 6). This satisfies the zero-shot criterion for the tested building population, though generalization to non-residential buildings remains untested.
    \item\emph{Cross-domain transfer:} We observe successful transfer across three distinct climate zones representing hot-humid (Texas, 2A), marine (California, 3C), and cold (Vermont, 6A) conditions. The cross-region model (trained on all three climates) achieves the best overall performance, suggesting that climate diversity in training improves generalization. Transfer to climates with substantially different characteristics (tropical, arid, subarctic) requires future validation.
    \item\emph{Sample efficiency:} Our most significant result is the data efficiency enabled by physics-informed design. Training on just 2-4 buildings approaches the performance of training on 32 buildings (Figure 2a). This represents approximately 8-16× reduction in data requirements compared to LSTM baselines, addressing a critical practical barrier to foundational model deployment.
\end{enumerate}

\subsection{Interpretation of Transfer Behavior}
The Physics Transformer's data efficiency stems from learning thermal response patterns (temperature change per unit input) rather than absolute ranges. This explains why 2–4 buildings in the training set are enough: once the model learns generalizable heat transfer dynamics, additional buildings provide diminishing returns.

California's weaker performance, both within-climate and as a transfer source, likely reflects its mild climate with limited thermal variability. Models trained on California data encounter fewer dynamic patterns, reducing their ability to generalize. Conversely, Texas's hot summers and variable seasons provide a richer training signal.

The 32-building Texas model's reduced transfer suggests a capacity limit: with sufficient data, the model may begin memorizing building-specific patterns rather than learning universal dynamics. This reinforces that data quality (dynamic range, seasonal diversity) matters more than quantity for transfer learning.

\subsection{Limitations} 
Several gaps remain before claiming a complete foundational model. Most significantly, our evaluation is limited to single-family residential buildings; commercial buildings with multi-zone HVAC systems, variable occupancy, and complex control sequences present substantially different dynamics that remain untested. Additionally, all training and evaluation uses EnergyPlus simulations rather than real building data. This gap may impact performance in real-world deployment and may require additional fine-tuning steps. Finally, our three U.S. climate zones (hot-humid, marine, and cold) represent some diversity but exclude extreme conditions such as tropical, and arid climates, as well as non-U.S. building stocks with different construction practices and operational norms.

\subsection{Future Directions}
We view this work as establishing necessary but not sufficient conditions for foundational thermal models: the architectural principles that enable transfer. Scaling these principles to broader building populations and validating on real-world data are essential next steps: Bridging the sim-to-real gap may require domain adaptation techniques, hybrid approaches that combine simulation pre-training with real-world fine-tuning, or uncertainty quantification methods that flag out-of-distribution inputs during deployment. Extending the approach to commercial buildings would test whether our physics-informed principles generalize beyond residential thermal dynamics to more complex systems with multiple zones and variable occupancy. Broader geographic coverage, across diverse international climate zones and building stocks, would establish true foundational scope and reveal climate-specific failure modes not apparent in our current evaluation.
\section{Conclusion}
\label{s:conclusion}
This paper takes concrete steps toward the vision of foundational thermal models for buildings. We introduced the Physics Transformer, demonstrating that physics-informed architectural principles, i.e, derivative enrichment and Euler-based residual prediction, enable zero-shot transfer across buildings and climates with minimal training data. Our key finding challenges the prevailing foundation model paradigm: for building thermal dynamics, domain-specific inductive biases outperform massive-scale generic architectures, even when the latter are fine-tuned on in-domain data. 

Experimental validation on a diverse, multi-climate dataset confirmed the generalization capabilities of the proposed model. The Physical Transformer consistently outperformed baseline LSTM and TimesFM models, demonstrating the ability to generalize to unseen buildings within a single climate zone. Notably, the model achieved state-of-the-art performance with significantly less training data than competing approaches, suggesting that the architecture captures the fundamental thermal dynamics rather than merely overfitting to dataset characteristics. Furthermore, the model showed strong cross-climate transferability, proving effective even when applied to regions outside its training distribution.
% with multi-region trained models proving robust and effective across all climate zones. This suggests the architecture successfully captures the fundamental, underlying thermal dynamics common to buildings, rather than merely overfitting to the characteristics of a specific dataset.

%Despite these promising results, this study identified clear limitations. While effective in one-step-ahead forecasting, the model's performance degraded in multi-step prediction scenarios, showing error accumulation and fluctuations around steady-state temperatures—a common challenge for data-driven models lacking hard physical constraints.

%Future work should focus on addressing these stability issues to enhance real-world applicability for control. Key directions include the integration of explicit energy constraints within the training objective and the exploration of energy-based model structures. 

%Several directions are essential for progressing from these foundational principles to a complete foundational model. First, validation on real building data must establish whether the sim-to-real gap can be bridged through domain adaptation or if real-world pre-training is required. Second, extending the approach to commercial buildings with multi-zone HVAC systems would test the universality of our physics-informed principles. Third, addressing multi-step prediction stability—through energy-based formulations or explicit equilibrium constraints—is necessary for control applications. Fourth, scaling to global climate zones and diverse building stocks would establish true foundational coverage.

We believe the architectural principles validated here—embedding heat transfer physics through derivative enrichment and residual prediction—constitute necessary ingredients for foundational building energy models. Our results provide empirical evidence that zero-shot transfer is achievable for thermal dynamics, establishing a proof-of-concept that motivates continued development toward models that are truly foundational: universal, transferable, and practically deployable across the global building stock.

To facilitate this research agenda, we release our code and trained models at [repository URL].

% talking about inductive bias

\bibliographystyle{ACM-Reference-Format} 
\bibliography{refs}

\end{document}